\documentclass[journal,10pt]{IEEEtran}
\usepackage{cite}
\usepackage{graphicx}
\usepackage{multirow}
\usepackage{color,soul}
\usepackage{tabularx}
\usepackage{tabu}
\usepackage{array}
\usepackage{stfloats}
\usepackage{enumitem}

\graphicspath{{./figures/}}
\usepackage[font=small,skip=2pt]{caption}

\begin{document}

\title{Object Detection Under Rainy Conditions for Autonomous Vehicles: A Review of State-of-the-Art and Emerging Techniques}


\author{Mazin Hnewa
        and Hayder Radha,~\IEEEmembership{Fellow,~IEEE}
\thanks{Mazin Hnewa, PhD student, Department of Electrical and Computer Engineering, Michigan State University, East Lansing, MI 48824 USA (email: mazin@msu.edu).}
\thanks{Hayder Radha, MSU Foundation Professor, Department of Electrical and Computer Engineering,Michigan State University, East Lansing, MI 48824 USA (email: radha@egr.msu.edu)}.
}

\maketitle

\vspace{-1cm}
\begin{abstract}
Advanced automotive active-safety systems, in general, and autonomous vehicles, in particular, rely heavily on visual data to classify and localize objects such as pedestrians, traffic signs and lights, and other nearby cars, to assist the corresponding vehicles maneuver safely in their environments. However, the performance of object detection methods could degrade rather significantly under challenging weather scenarios including rainy conditions. Despite major advancements in the development of deraining approaches, the impact of rain on object detection has largely been understudied, especially in the context of autonomous driving. The main objective of this paper is to present a tutorial on state-of-the-art and emerging techniques that represent leading candidates for mitigating the influence of rainy conditions on an autonomous vehicle\textsc{'}s ability to detect objects. Our goal includes surveying and analyzing the performance of object detection methods trained and tested using visual data captured under clear and rainy conditions. Moreover, we survey and evaluate the efficacy and limitations of leading deraining approaches, deep-learning based domain adaptation, and image translation frameworks that are being considered for addressing the problem of object detection under rainy conditions. Experimental results of a variety of the surveyed techniques are presented as part of this tutorial.
\end{abstract}


\IEEEpeerreviewmaketitle

\section{Introduction}

\IEEEPARstart{V}{isual} data plays a critical role in enabling automotive advanced driver-assistance systems and autonomous vehicles to achieve high levels of safety while the cars and trucks maneuver in their environments. Hence, emerging autonomous vehicles are employing cameras and deep learning-based methods for object detection and classification \cite{chen2015deepdriving, wu2017squeezedet,teichmann2018multinet}. These methods predict bounding boxes that surround detected objects and classify probabilities associated with each bounding box. In particular, convolutional neural network (CNN)-based approaches have shown very promising results in the detection of pedestrians, vehicles, and other objects \cite{girshick2014rich,girshick2015fast,FasterRCNN,RetinaNet,YOLO,liu2016ssd,R-FCN}. These neural networks are usually trained using a large amount of visual data captured in favorable clear conditions. However, the performance of such systems in challenging weather, such as rainy conditions, has not been thoroughly surveyed or studied.

The quality of visual signals captured by autonomous vehicles can be impaired and distorted in adverse weather conditions, most notably in rain, snow, and fog. Such conditions minimize the scene contrast and visibility, and this could lead to a significant degradation in the ability of the vehicle to detect critical objects in the environment. Depending on the visual effect, adverse weather conditions can be classified as steady (such as fog, mist, and haze) or dynamic, which have more complex effects (such as rain and snow) \cite{garg2004detection}. In this article, we focus on rain because it is the most common dynamic challenging weather condition that impacts virtually every populated region of the globe. Furthermore, there has been a great number of recent efforts that attempt to mitigate the effect of rain in the context of visual processing. While addressing the effect of other weather conditions has been receiving some, yet minimal, attention, the volume of work regarding the mitigation of rain is far more prevalent and salient within different research communities.

It is worth noting that rain consists of countless drops that have a wide range of sizes and complex shapes, and rain spreads quite randomly, with varying speeds when falling on roadways, pavement, vehicles, pedestrians, and other objects in the scene. Moreover, raindrops naturally cause intensity variations in images and video frames. In particular, every raindrop blocks some of the light that is reflected by objects in a scene. In addition, rain streaks lead to low contrast and elevated levels of whiteness in visual data. Consequently, mitigating the effect of rain on visual data is arguably one of the most challenging tasks that autonomous vehicles will have to perform, due to the fact that it is quite difficult to detect and isolate raindrops, and it is equally problematic to restore the information that is lost or occluded by rain.  

Meanwhile, there has been noticeable progress in the development of advanced visual deraining algorithms \cite{6099619,yang2017deep,DNN,DeRaindrop, PReNet,wang2019spatial}. 
Thus, one natural and intuitive solution for mitigating the effect of rain on active safety systems and autonomous vehicles is to employ robust deraining algorithms and then apply the desired object detection approach to the resulting derained signal. State-of-the-art deraining algorithms, however, are designed to remove the visual impairments caused by rain, while attempting to restore the original signal with minimal distortion. Hence, the primary objective of these algorithms, in general, is to preserve the visual quality as measured by popular performance metrics, such as the peak signal-to-noise-ratio and structure similarity index (SSIM)\cite{SSIM}. These metrics, however, do not reflect a viable measure for analyzing the performance of the system for more complex tasks, such as object detection.

The main objective of this article is to survey and present a tutorial on state-of-the-art and emerging techniques that are leading candidates for mitigating the influence of rainy conditions on an autonomous vehicle’s ability to detect objects. In that context, our goal includes surveying and analyzing the performance of object detection methods that are representatives of state-of-the-art frameworks that are being considered for integration into autonomous vehicles’ artificial intelligence (AI) platforms. Furthermore, we survey and highlight the inherent limitations of leading deraining algorithms, deep learning-based domain adaptation, and image translation frameworks in the context of rainy conditions.

\begin{figure*}[!t]
\centering
\includegraphics[width=1\linewidth]{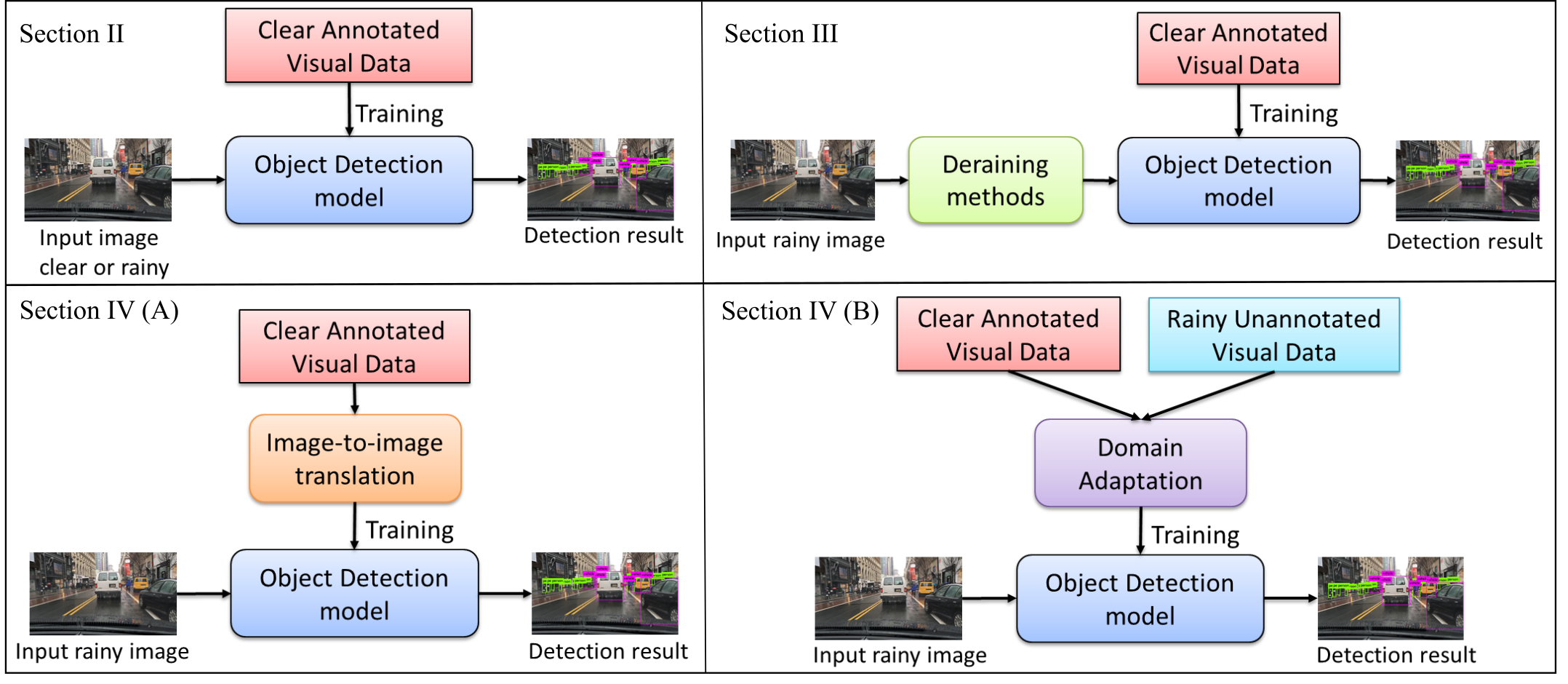}
\caption{The architectures highlighting the main parts of the article. The (a) “Object Detection for Autonomous Vehicles in Clear and Rainy Conditions” section, (b) “Deraining in Conjunction With Object Detection” section, (c) “Unsupervised Image-to-Image Translation” section, and (d) “Domain Adaptation” section.}
\label{Fig_main}
\vspace{-0.6cm}
\end{figure*}

While surveying a variety of relevant techniques in this area, we present experimental results with the objective of highlighting the urgent need for developing new paradigms for addressing the challenges of autonomous driving in severe weather conditions. Although generative model-based image translation and domain adaptation approaches do show some promise, one overarching conclusion that we aim to convey through this article is that current solutions do not adequately mitigate the realistic challenges for autonomous driving in diverse weather conditions. This overarching conclusion opens the door for the research community to pursue and explore new frameworks that address this timely and crucial problem area. The architectures highlighting the main parts of this article are highlighted in Figure \ref{Fig_main}.

\section{Object detection for autonomous vehicles in clear and rainy conditions} \label{sec_rain_impact}
The level of degradation in the performance of an object detection method, trained in certain conditions, is influenced heavily by 1) how different the training and testing domains are and 2) the type of deep learning-based architecture used for object detection. Most recent object detectors are CNN-based networks, such as the single-shot multibox detector \cite{liu2016ssd}, region-based fully convolutional network \cite{R-FCN}, You Only Look Once (YOLO) \cite{YOLO}, RetinaNet \cite{RetinaNet}, and Faster R-CNN\cite{FasterRCNN}. To that end, we review two major classes of object detection frameworks that are both popular and representative of deep learning-based approaches. As we see later in this tutorial, these two classes of architectures exhibit different levels and forms of degradation in response to challenging rainy conditions, and they also perform rather differently in conjunction with potential rain mitigation frameworks.

In particular, we briefly describe the underlying architectures for Faster R-CNN and YOLO as representatives of two major classes of object detection algorithms. Faster R-CNN is arguably the most popular of the object detection algorithms that are based on a two-stage deep learning architecture; one stage is for identifying region proposals (RPs), and the second is for refining and assigning class probabilities for the corresponding regions. YOLO, on the other hand, is a representative of detection frameworks that directly operate on the whole image.

\subsection{Deep learning-based methods for object detection}

The utility of CNNs for object detection was well established prior to the introduction of the notion of RPs, commonly known as R-CNN \cite{girshick2014rich},  where “R” stands for regions or RPs. A fast version of R-CNN was later introduced \cite{girshick2015fast}, and then Ren \textit{et al.} \cite{FasterRCNN} presented the idea of the RP network (RPN) that shares convolutional layers with Fast R-CNN \cite{girshick2015fast}. The RPN is merged with the Fast R-CNN into one unified network that is known as Faster R-CNN to achieve more computationally efficient detection. Under Faster R-CNN, an input image is fed to a feature extractor, such as the ZF model \cite{ZF} or VGG-16 \cite{VGG16}, to produce a feature map. Then, the RPN utilizes this feature map to predict RPs (regions in the image that could potentially contain objects of interest).

\begin{figure*}[!t]
\centering
\includegraphics[width=0.95\linewidth]{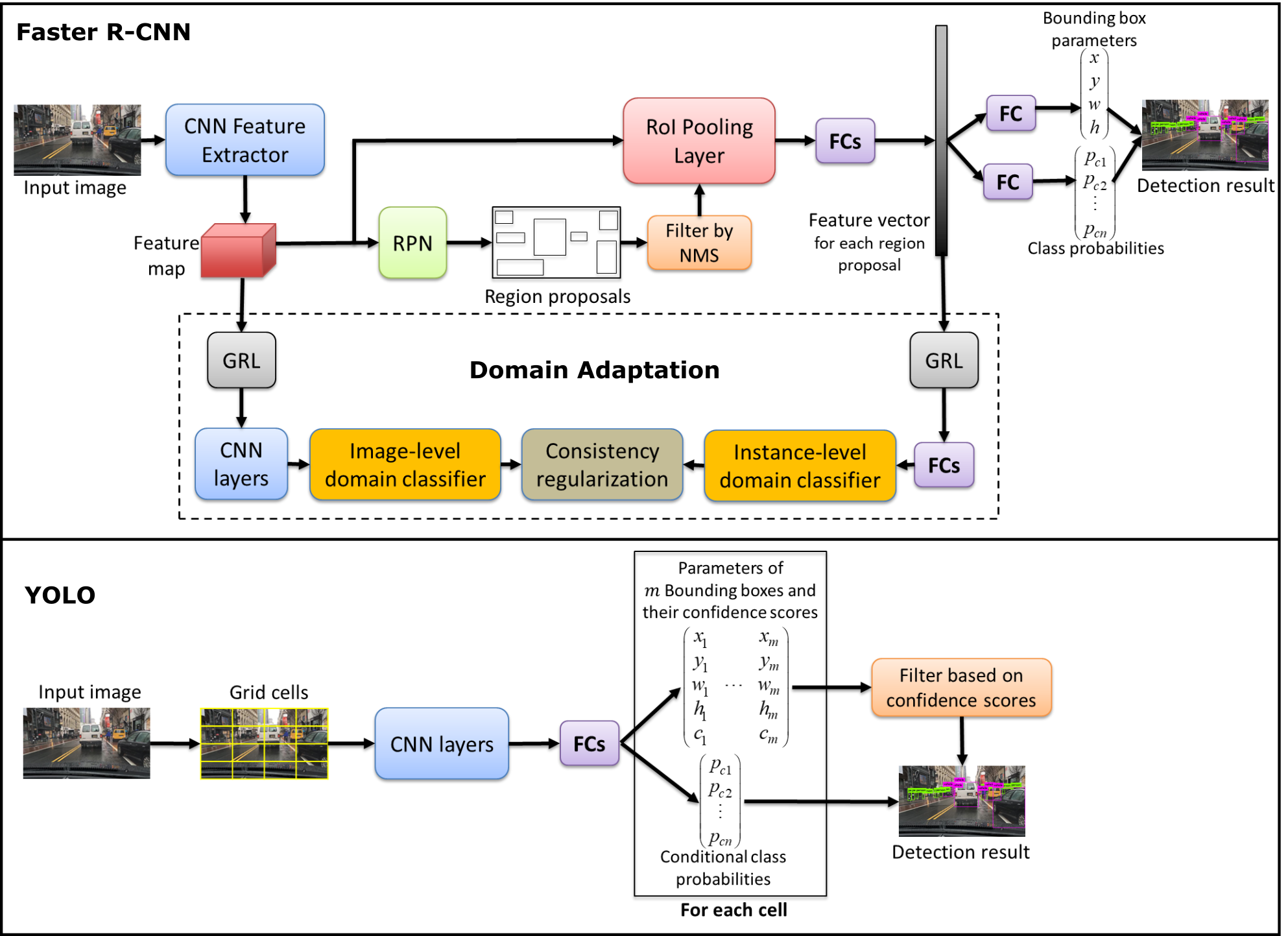}
\caption{The high-level architectures of the detection methods that are used in this tutorial. The domain adaptation of Faster R-CNN is explained in Section \ref{DA}. GRL: gradient reversal layer. .}
\label{Fig_faster-RCNN}
\vspace{-0.6cm}
\end{figure*}

In that context, many RPs are quite overlapped with each other, with significant numbers of pixels common among multiple RPs. To filter out the substantial redundancy that might occur with such a framework, nonmaximum suppression (NMS) \cite{felzenszwalb2009object} is used to remove redundant regions while keeping the ones that have the highest prediction scores. Subsequently, each regional proposal that survives the NMS process is used by a region-of-interest (RoI) pooling layer to crop the corresponding features from the feature map. This cropping process produces a feature vector that is fed to two fully connected layers: one predicts the offset values of a bounding box of an object with respect to the regional proposal, and the other predicts class probabilities for the predicted bounding box. Figure \ref{Fig_faster-RCNN} shows a high-level architecture of Faster R-CNN.

On the other hand, Redmon \textit{et at.} \cite{YOLO} proposed to treat object detection as a regression problem, and they developed a unified neural network that is called \textit{YOLO} to predict bounding boxes and class probabilities directly from a full image in one evaluation. Under YOLO, an input image is divided into a specific set of grid cells, and each cell is responsible for detecting objects whose centers are located within that cell. To that end, each cell predicts a certain number of bounding boxes, and it also predicts the confidence scores for these boxes in terms of the likelihood that they contain an object. Furthermore, it predicts conditional class probabilities, given that it has an object. In this case, there are potentially many wrongly predicted bounding boxes. To filter them out and provide the final detection result, a threshold is used on the confidence scores of the predicted bounding boxes. Figure \ref{Fig_faster-RCNN} illustrates the general architecture of YOLO.

\subsection{Object detection performance for neural network architectures in clear and rainy conditions} \label{sec_dataset}
Here, we provide an insight into the level of degradation caused by rainy conditions on the performance of the two major deep learning architectures described previously. In particular, we focus on the following fundamental question: how much degradation a deep neural network that is trained in clear conditions will suffer when tested in rainy weather. In that context, we first describe the data set that we used for training and testing; this is followed by presenting some visual and numerical results. For the purpose of this tutorial, we needed a rich data set that was captured in diverse weather conditions. Despite the fact that there are few notable data sets  \cite{kitti,Cityscapes,MVD2017}, which are quite popular among the computer vision and AI research communities in terms of training deep NNs, there is only one (arguably two \cite{yu2018bdd100k}\cite{nuscenes2019}) that is properly labeled and annotated for our purpose and hence that could be used for training and testing for different weather conditions. In particular, we use the Berkeley Deep Drive (BDD100K) data set \cite{yu2018bdd100k} because it contains image tagging for weather (i.e., each image in the data set is labeled with its weather condition, such as clear, rainy, foggy, and so on). Meanwhile, although some other data sets, such as nuScenes \cite{nuscenes2019}, might contain some visuals captured in rainy conditions, they do not have weather tagging. Hence, choosing the BDD100k data set was influenced by the fact that we could select images illustrating a specific weather condition. 

\begin{figure*}[!t]
\centering
\includegraphics[width=1\linewidth]{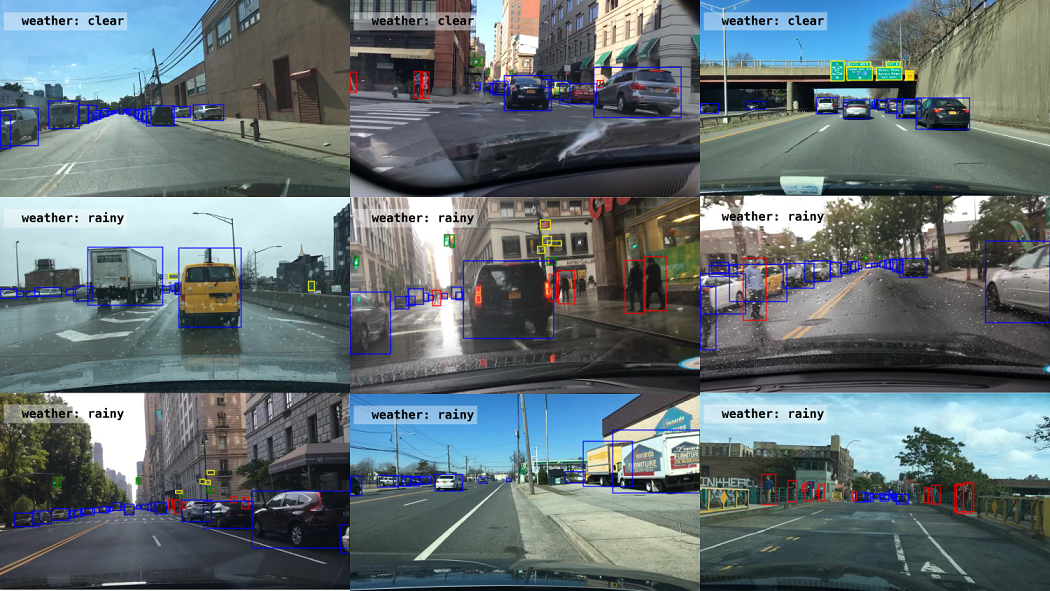}
\caption{The examples of the annotated images in the BDD100K data set \cite{yu2018bdd100k}. Images in the top row are tagged as \textit{clear weather}, and images in the middle and bottom rows are tagged as being captured in \textit{rainy weather}. However, images in the bottom row are wrongly tagged as being rainy weather, but they actually show in clear or cloudy weather.}
\label{Fig1}
\end{figure*}

Moreover, the BDD100K has 100,000 video clips captured in diverse geographic, environmental, and weather conditions. It is worth noting that only one selected frame from each video is annotated with object bounding boxes as well as image-level tagging. Examples of annotated frames in clear and rainy weather are shown in Figure \ref{Fig1}. In this article, we consider the four classes (vehicle, pedestrian, traffic light, and traffic sign) that are labeled and provided as ground-truth objects within the BDD100K data set. Naturally, these four classes are among the most critical objects for an autonomous vehicle. In this tutorial, we use images that are captured in clear weather from the designated training set of the BDD100K to form our underlying training data set. We refer to this training data as the \textit{train clear} set, which we used consistently to train the detection methods for the different scenarios covered in this article. For testing, we use a collection of clear weather images from the testing set of the BDD100K. We refer to this latter group as the \textit{test clear} set. Table \ref{data_set} gives the number of annotated objects in the train clear and test clear data sets.   

 One approach to demonstrate the impact of rain on object detection methods that are trained in clear conditions is by rendering synthetic rain \cite{realistic_rain_rendering,photorealistic,Adobe_after} within the images of the test clear set. Then, the synthetic rainy data can be used to test the already trained object detection methods. The benefit of this approach is that one would have the exact same underlying content in both testing data sets in terms of the objects within the scene, with one set representing the original clear-weather content when the data was captured and another set with the synthetic rain. This would plainly show the impact of rain, as the visual objects are the same in both tested sets (the test clear set and a test synthetic rain set). However, from our extensive experience in this area, we noticed that most well-known rain simulation methods do not render realistic rain that viably captures actual and true rainy weather conditions, especially for a driving vehicle. Thus, when comparing the two scenarios, this discrepancy between synthetic and natural (real) rainy conditions will lead to domain mismatch. As a result, we do not test the detection methods using synthetic rain in our study because those techniques will not demonstrate the impact of true natural rain on a driving vehicle.

Alternatively, we use images captured in real rainy conditions from the training and testing sets of the BDD100K data set to test the object detection methods. It is worth noting that several images in the data set are wrongly tagged as rainy weather when they actually show clear or cloudy conditions, such as the examples shown in the bottom row of Figure \ref{Fig1}. To solve this problem, we manually selected the images that were truly captured in rainy weather to form what we refer to as the \textit{test rainy} set. Equally important, we elected to have both the test clear and test rainy sets include approximately the same number of annotated objects, as shown in Table \ref{data_set} in order to provide statistically comparable results. 

\begin{table}[!t]
\centering
\caption{The number of annotated objects in  training and testing sets that are used in our study.}
\label{data_set}
\begin{tabular}{|c|c|c|c|c|}
\hline
\textbf{Set} &  \textbf{Vehicles} & \textbf{Pedestrians}& \textbf{Traffic signs} & \textbf{Traffic lights}\\
\hline
Train clear  & 149,548 & 16777 & 43866 & 26002  \\
\hline
Test clear & 13721 & 2397 & 3548 & 4239\\
\hline
Test rainy  & 13724  & 2347 & 3551 & 4246\\
\hline
\end{tabular}
\end{table}

It is important to make one final critical note regarding the currently available data sets for training NNs designed for object detection. The lack of data sets captured in diverse conditions, including rain, snow, fog, and other weather scenarios, represents one of the most challenging aspects of achieving a viable level of training for autonomous vehicles. Even for the BDD100K data set, which is one of very few publicly available data sets with properly annotated objects captured in different weather conditions, there is not a sufficient amount of annotated visual content that is truly viable for training in rainy weather. This fundamental issue with the lack of real training data for rainy and other conditions has clearly become a major obstacle, to the extent that leading high-tech companies working in the area have begun a focused effort designated specifically for collecting data in rainy conditions. For example, Waymo recently announced plans to begin collecting data for autonomous driving in rainy conditions\cite{Waymo-Florida}.  

\subsection{Performance metric}
To evaluate the detection performance, we compute the mean average precision (mAP). This metric has been the most popular performance measure since the time when it was originally defined in the PASCAL Visual Object Classes Challenge 2012 for evaluating detection methods \cite{2011pascal}. To determine the mAP, a precision/recall curve is first computed based on the prediction result against the ground truth. A prediction is considered a true positive if its bounding box has 1) an intersection-over-union value greater than 0.5 relative to the corresponding ground-truth bounding box and 2) the same class label as the ground truth. Then, the curve is updated by making the precision monotonically decrease. This is achieved by setting the precision for recall $r$ to the maximum precision obtained for any recall $r'>r$. The AP is the area under the updated precision/recall curve. It is computed by numerical integration. Finally, the mAP is the mean of the AP among all classes.

\subsection{Results and Discussion} \label{1_results}
\begin{figure*}[!t]
\centering
\includegraphics[width=1\linewidth]{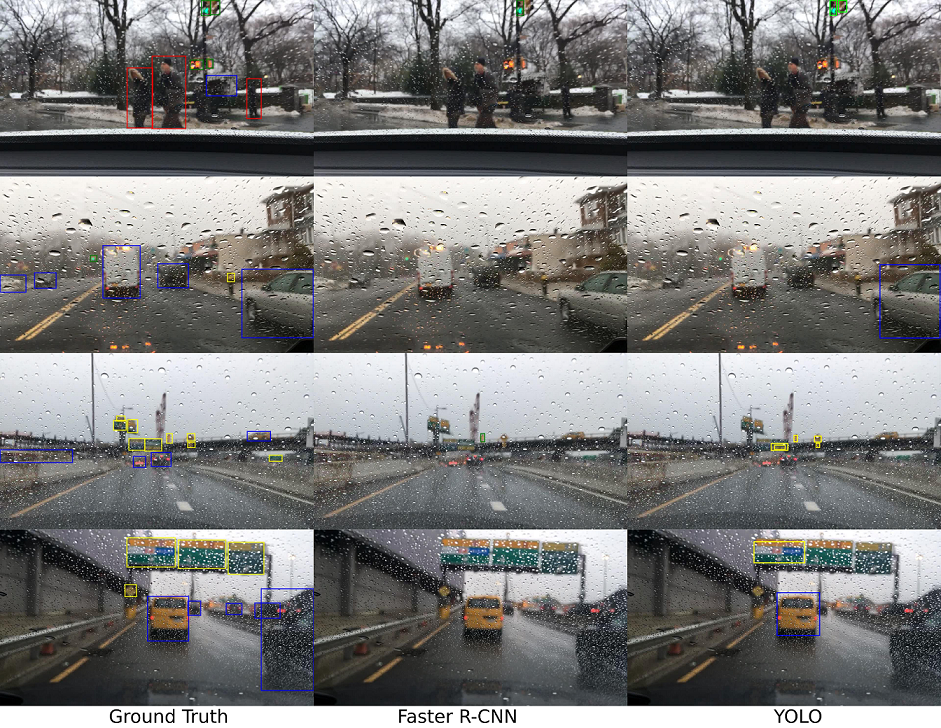}
\caption{Examples of detection results using Faster R-CNN and YOLO for different visual scenes from \textit{test rainy set}}
\label{Fig_res_detection}
\vspace{-0.6cm}
\end{figure*}

We trained the detection methods (Faster R-CNN and YOLO) using the train clear set, which is described in section \ref{sec_dataset}. We used the same training settings and hyper-parameters that were used in the original papers \cite{FasterRCNN}\cite{YOLO}. Then, we tested the trained models using the test clear and test rainy sets to illustrate the impact of rain. Table \ref{mAP} presents the AP for each class as well as the mAP evaluated based on the AP values of the classes. From the table, we observe that the mAP clearly declines in rainy weather compared to clear weather using both Faster R-CNN and YOLO. Consequently, these results undoubtedly illustrate that the performance of an object detection framework that is trained using clear visuals could significantly degrade in rainy weather conditions. The performance decreases due to the fact that rain covers and distorts important details of the underlying visual features, which are used by detection methods to classify and localize objects. Figure \ref{Fig_res_detection} provides examples when the detection methods fail to perceive most objects in rainy conditions. 

\begin{table*}[!t]
\centering
\caption{Average precision (AP) for each class, and  mean average precision (mAP) evaluated based on the AP values of the classes. V-AP: vehicle, P-AP: pedestrian, TL-AP: traffic light, and TS-AP: traffic sign average precision. *The top row shows the performance under clear conditions (i.e., using the \textit{test clear} set), while all other rows show the performance under rainy conditions (i.e., using the \textit{test rainy} set). **Significant degradation in performance can be observed due to rainy conditions (text in red) relative to the performance under clear conditions (top row). Improvements in performance by mitigating the effect of rain can be observed using generative model-based image translation and/or domain adaptation (highlighted in bold). Meanwhile, deraining algorithms do not improve, and most of the time further degrade the performance.}
\label{mAP}
\begin{tabular} {|c|c|c|c|c|c|c|c|c|c|c|}
\hline
\multirow{2}{*}{\textbf{Mitigating technique}} & \multicolumn{5}{c|}{\textbf{Faster R-CNN}}  & \multicolumn{5}{c|}{\textbf{YOLO-V3}} \\\cline{2-11}
  & \textbf{V-AP} & \textbf{P-AP} & \textbf{TL-AP} & \textbf{TS-AP} &\textbf{mAP} & \textbf{V-AP} & \textbf{P-AP} & \textbf{TL-AP} & \textbf{TS-AP} & \textbf{mAP}   \\
\hline
None (clear conditions*) & 72.61 & 40.99 & 26.07 & 38.12 & 44.45 & 76.57 & 37.12 & 46.22 & 50.56 & 52.62 \\
\hline
\hline
None (rainy conditions**) &  \textcolor{red}{67.84} & \textcolor{red}{32.58} & \textcolor{red}{20.52} & \textcolor{red}{35.04} & \textcolor{red}{39.00} & \textcolor{red}{74.15} & \textcolor{red}{32.07} & \textcolor{red}{41.07} & \textcolor{red}{50.27} & \textcolor{red}{49.39}  \\
		\hline
Deraining DDN \cite{DNN} &  67.00 & 28.55 &  20.02  & 35.55 & 37.78 & 73.07 & 29.89 & 40.05 & 48.74 & 47.94 \\
\hline
Deraining DeRaindrop \cite{DeRaindrop} &  64.37 & 29.27 & 18.32  & 33.33 & 36.32 & 70.77 & 30.16 & 37.70 & 48.03 & 46.66  \\
\hline
Deraining PReNet \cite{PReNet} &  63.69 & 24.39 & 17.40  & 31.68 & 34.29 & 70.83 & 27.36 & 35.49 & 43.78 & 44.36 \\
\hline
Image translation UNIT \cite{unit} &  \textbf{68.47} & \textbf{32.76} & 18.85 & \textbf{36.20} & \textbf{39.07}
 & 74.14 & \textbf{34.19} & \textbf{41.18} & 48.41 & \textbf{49.48}  \\
\hline
Domain adaptation \cite{chen2018domain} & 67.36 & \textbf{34.89} & 19.24 & \textbf{35.49} & \textbf{39.24} & \multicolumn{5}{c|}{Not applicable}  \\
\hline
\end{tabular}
\vspace{-0.3cm}
\end{table*}

Moreover, one can notice that in rainy conditions, the AP for the pedestrian and traffic light classes declines more significantly than the decrease in performance for vehicle and traffic sign classes. This discrepancy in performance degradation for different objects is due to a variety of factors. For example, vehicles usually occupy larger regions within an image frame than other types of objects; hence, even when raindrops or rain streaks cover a visual of a vehicle, there are still sufficient features that can be extracted by the detection method. Furthermore, traffic signs are normally made from materials that have high reflectivity, which makes it easier for an object detection method to achieve higher accuracy, even when a traffic sign visual is distorted by some rain. Overall, in both cases, the important features needed for reliable detection are still salient within the underlying deep NNs of the detection algorithms. Nevertheless, rain could still impact the detection of vehicle and traffic signs, as shown in the bottom three rows of Figure \ref{Fig_res_detection}.

\section{Deraining in conjunction with object detection} \label{sec_deraining}
\begin{figure*}[!t]
\centering
\includegraphics[width=1\linewidth]{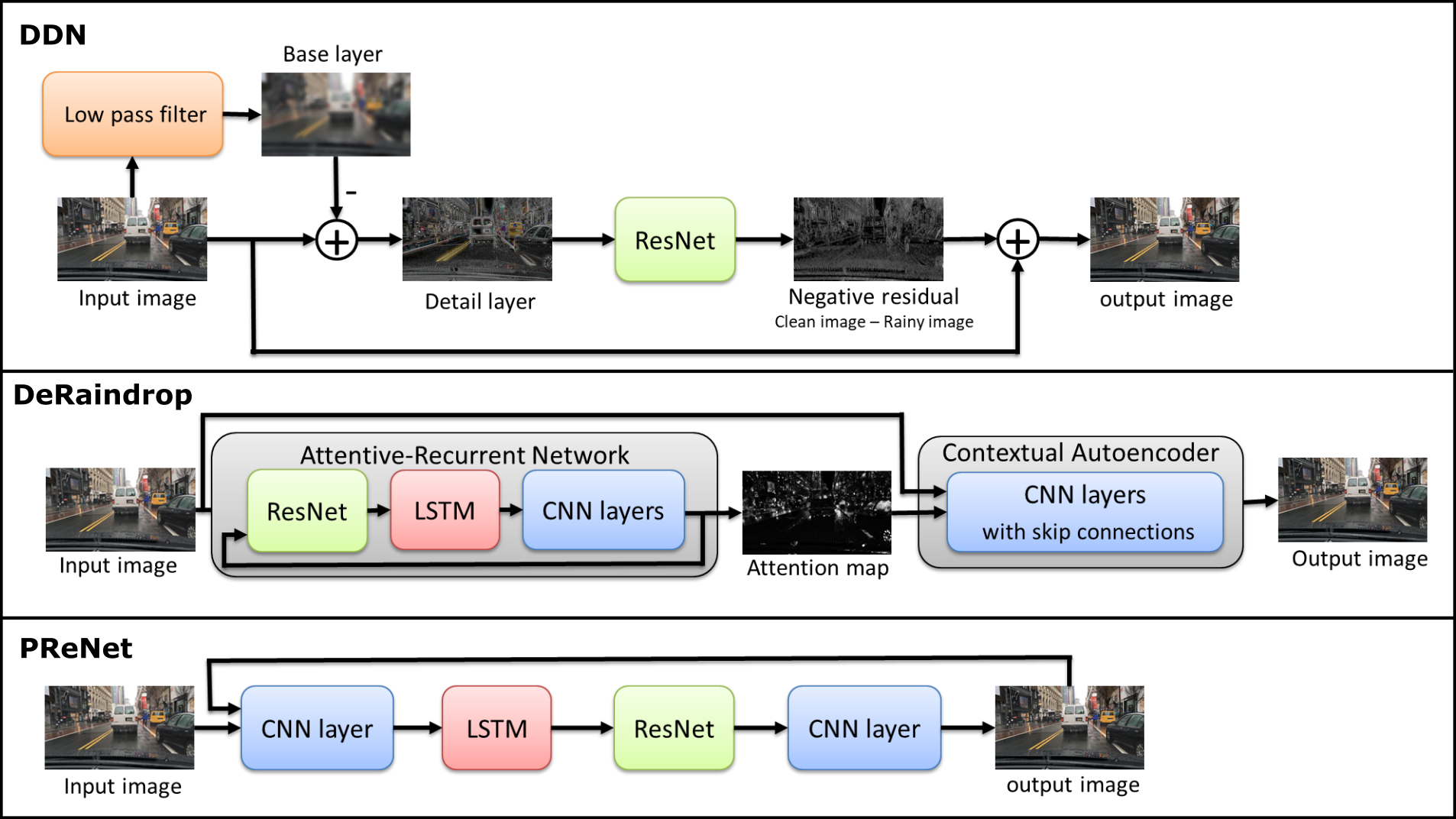}
\caption{ The General architectures of the used deraining methods (DDN \cite{DNN}, DeRaindrop \cite{DeRaindrop}, and PReNet \cite{PReNet}). LSTM: long short-term memory; ARN: attentive recurrent network. }
\label{Arc_deraining}
\vspace{-0.6cm}
\end{figure*}

Deraining methods attempt to remove the effect of rain and restore an image of a scene that has been distorted by raindrops or rain streaks while preserving important visual details. In this tutorial, we review three recently developed deraining algorithms \cite{DNN,DeRaindrop, PReNet} that employ deep learning frameworks for the removal of rain from a scene. The high-level architectures of these methods are shown in Figure \ref{Arc_deraining}. Below, we briefly describe these three deraining methods and highlight their limitations when employing them in conjunction with object detection methods.

\subsection {Deep Detail Network}
Fu \textit{et at.} \cite{DNN} proposed a Deep Detail Network (DDN) to remove rain from a single image. They employed a convolutional neural network (CNN), which is ResNet \cite{ResNet} to predict the difference between clear and rainy images, and use this difference to remove rain from a scene. In particular, the DDN exploits the rainy image's high frequency details only, and it uses such details as input to ResNet while ignoring the low frequency background (interference) of the same underlying scene.

\subsection{Attentive Generative Adversarial Network}
Qian \textit{et at.} \cite{DeRaindrop} proposed attentive generative adversarial network that is called "DeRaindrop" to remove raindrops from images. In this method, a generative adversarial network (GAN) \cite{GAN} with visual attention is employed to learn raindrop areas and their surroundings. The first part of the generative network, known as the Attentive-Recurrent Network (ARN), produces an attention map to guide the next stage of the DeRaindrop framework. ARN includes ResNet, a Long Short-Term Memory (LSTM) \cite{LSTM}, and CNN layers. The second stage of DeRaindrop, which is known as \textit{Contextual Autoencoder}, operates on the attention map and hence it focuses on (or "pay more attention" to) the raindrop areas. The overall process from the two stages is expected to clean images free of raindrops. The architecture also includes a discriminative network, which assesses the generated rain-free images to verify that they are similar to real ones that have been used during the training process.    

\subsection{Progressive Image Deraining Network}
Ren \textit{et at.} \cite{PReNet} proposed a Progressive Recurrent Network (PReNet) to recursively remove rain from a single image. At each iteration, some rain is removed, and the remaining rain can be progressively removed in subsequent iterations. Consequently, after a certain number of iterations, most of the rain should be removed leading to a rain-free quality image. In addition to several residual blocks of ResNet, PReNet includes a CNN layer that operates on both the original rainy image and current output image. PReNet also includes another CNN layer to generate the output image. Furthermore, a recurrent layer is appended to exploit dependencies of deep features across iterations via convolutional LSTM. To train PReNet, a single negative SSIM \cite{SSIM} or MSE loss is used.

\subsection{Results and Discussion}
To demonstrate the performance of the deraining methods outlined previously, we apply the pretrained deraining models provided by the corresponding authors to the test rainy set as a prepossessing step. After applying the deraining algorithms, which are anticipated to remove the rain from the input visual data and generate rain-free clear visuals, we feed the derained images into the object detection methods. Table \ref{mAP} shows the performance of the detection methods after applying the deraining approaches. It can be seen that the deraining algorithms actually degrade the detection performance when compared to directly using the rainy images as input into the corresponding detection frameworks. This is true for both Faster R-CNN and YOLO. One important factor for this degradation in performance is that the deraining process tends to smooth out the input image, and hence it distorts the meaningful information and distinctive features of a scene while attempting to remove the effect of rain. 

\begin{figure*}[!t]
\centering
\includegraphics[width=1\linewidth]{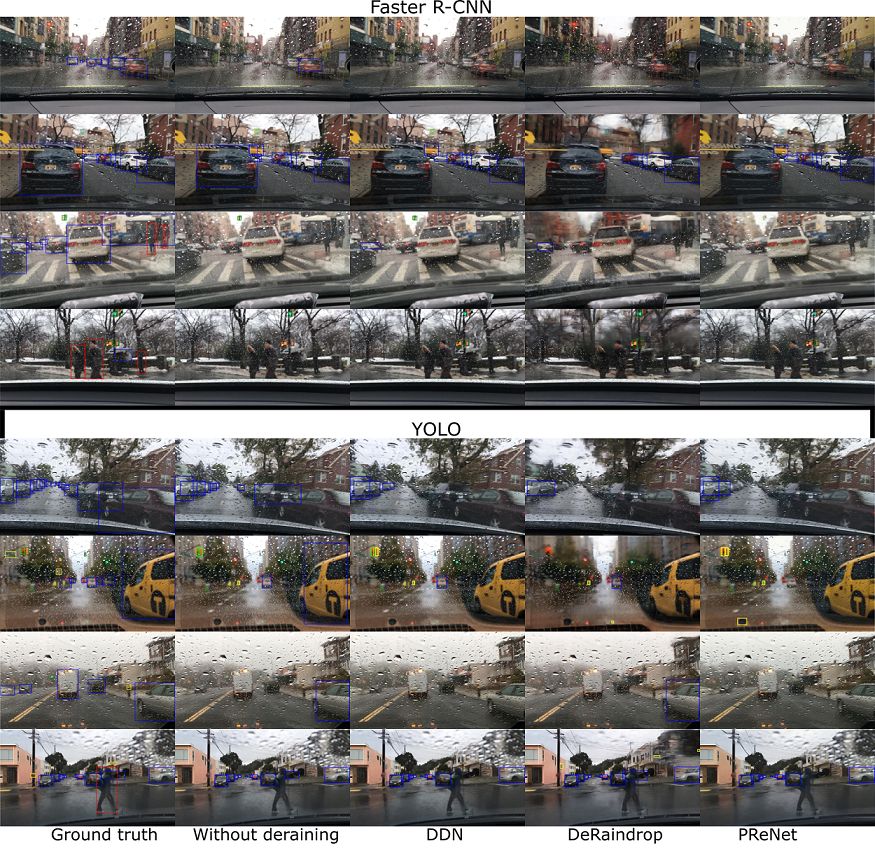}
\caption{he example detection results for different visual scenes where no deraining methods were employed and where deraining methods (DDN \cite{DNN}, DeRaindrop \cite{DeRaindrop}, and PReNet \cite{PReNet}) were used in conjunction with detection methods. Objects were detected using Faster R-CNN \cite{FasterRCNN} in the first group of images, and YOLO \cite{redmon2018yolov3} in the second group.}
\label{Fig6_derain}
\vspace{-0.4cm}
\end{figure*}

In particular, it is rather easy to observe that state-of-the-art deraining algorithms smooth out the edges of objects in an image, which leads to a loss of critical information and features, which are essential for enabling the detection algorithms to classify and localize objects. The images in the top two rows of \ref{Fig6_derain}, representing outputs of Faster R-CNN and YOLO, show some of the objects that are not detected after using the deraining methods but that are successfully detected if rainy images are directly used as input into the detection algorithms.

A related critical issue to highlight for current deraining algorithms is their inability to remove natural raindrops found in realistic scenes captured by moving vehicles. The root cause of this issue is the fact that deraining algorithms have been largely designed and tested using synthetic rain visuals superimposed on the underlying scenes. What aggravates this issue is that, at least in some cases, the background environments employed to design and test deraining algorithms are predominantly static scenes with a minimal number of moving objects. Consequently, the salient differences between such synthetic scenarios and the realistic environment encountered by a vehicle that is moving through natural rain represents a domain mismatch that is too much to handle for current deraining algorithms, and this leads to the algorithms’ failure under realistic conditions for autonomous vehicles. Hence, overall, we believe that relying purely on state-of-the-art deraining solutions does not represent a viable approach for mitigating the impact of rain on object detection. The images shown in Figure \ref{Fig6_derain}, especially some of the cases in the bottom two rows, illustrate examples of the failure of deraining methods to improve the performance of object detection.

\section{Alternative training approaches for deep learning based object detection} \label{sec_alt}
The requirement that autonomous driving systems must work reliably in different weather conditions is at odds with the fact that the training data are usually collected in dry weather with good visibility. Thus, the performance of object detection algorithms degrades in challenging weather conditions, as we showed in Section \ref{1_results}.

One simple approach to address this problem is to train a given CNN for the detection of objects using images captured in real rainy weather. As we highlighted earlier, sufficient annotated data sets captured by moving vehicles in realistic urban environments in natural rainy conditions are not readily available. To that end, and in spite of the fact that some data sets are available, the very few data sets captured under real rainy conditions are not properly annotated \cite{yu2018bdd100k}. Having such small data sets inherently makes them inadequate to reliably train deep learning architectures for objection detection. Furthermore, annotating the available data captured in real rainy conditions with accurate bounding boxes is an expensive and time-consuming process.

An alternative approach for addressing the lack-of-real-data issue is to train detection methods using synthetic rain data. However, and as we highlighted earlier, the trained methods generalize poorly on real data due to the domain shift between synthetic and natural rain. To solve this issue, we review approaches that can be employed for training the detection methods using annotated clear data in conjunction with unannotated rainy data. In particular, we review and survey two emerging frameworks for addressing this critical issue: image translation and domain adaptation.

\subsection{Unsupervised image-to-image translation}
Image-to-image translation (I2IT) is a well-known computer vision framework that translates images from one domain (e.g., captured in clear weather) to another domain (e.g., rainy conditions) while preserving the underlying and critical visual content of the input images. In short, I2IT attempts to learn a joint distribution of images in different domains. The learning process can be classified into a supervised setting, where the training data set consists of paired examples of the same scene captured in both domains (e.g., clear and rainy conditions), and an unsupervised setting, where different examples of both domains are used for training; hence, these examples do not have to be taken from the same corresponding scene.

The unsupervised case is inherently more challenging than supervised learning. More importantly, to address the main issue we face in the context of the lack of data needed to train object detection architectures in realistic conditions, we consequently need an unsupervised setting. In particular, the requirement of having a very large set of image pairs, where each pair of images must be of the same scene captured in different domains, renders supervised I2IT solutions virtually useless for our purpose. In fact, this requirement imposes more constraints than the lack-of-data issue that we are already trying to address. Hence, and despite the availability of well-known supervised learning-based techniques in this area \cite{pix2pix}\cite{SRGAN}, we have to resort to unsupervised solutions to address the problem at hand.

Recently, Generative Adversarial Networks (GANs) \cite{GAN} have been achieving very promising performance results in the area of image translation \cite{unit,pix2pix,SRGAN,cycleGan,bicycleGAN}. In general, a GAN consists of a \textit{generator} and a \textit{discriminator}. The generator is trained to \textit{fool} the discriminator, while the later is attempting to distinguish (or discriminate) between real natural images on one hand and fake images, which are generated by the trained generator, on the other hand. By doing this, GANs align the distribution of translated images with real images in the target domain.

\begin{figure*}[!t]
\centering
\includegraphics[width=1\linewidth]{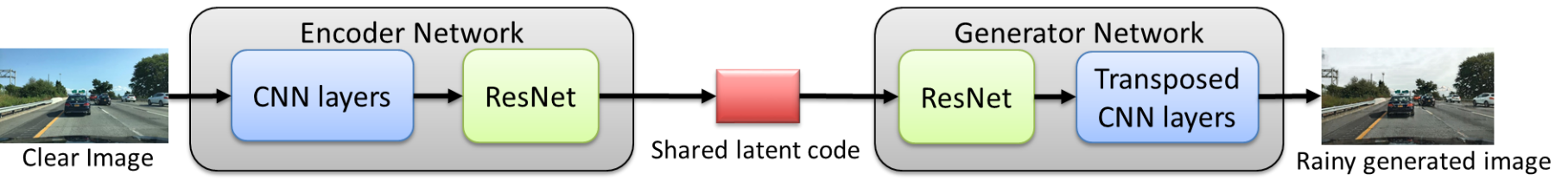}
\caption{The high-level architecture of the UNIT model \cite{unit} to generate images.}
\label{Fig3}
\vspace{-0.3cm}
\end{figure*}

As mentioned above, data sets that have paired clear-rainy images in driving environment is not publicly available. As a result, we use UNsupervised Image-to-image Translation (UNIT) \cite{unit} to translate clear images to rainy ones since the training process for the UNIT framework does not require paired images of the same scene. In other words, training UNIT requires two independent sets of images where one consists of images in one domain, and another set consists of images in another domain. The high-level architecture of the UNIT model is shown in Figure \ref{Fig3}. First, the encoder network maps an input image to a shared latent code (a shared compact representation of an image in both domains). Then, the generator network uses the shared latent code to generate an image in the desired domain. 

\begin{figure}[!t]
\centering

\includegraphics[width=1\linewidth]{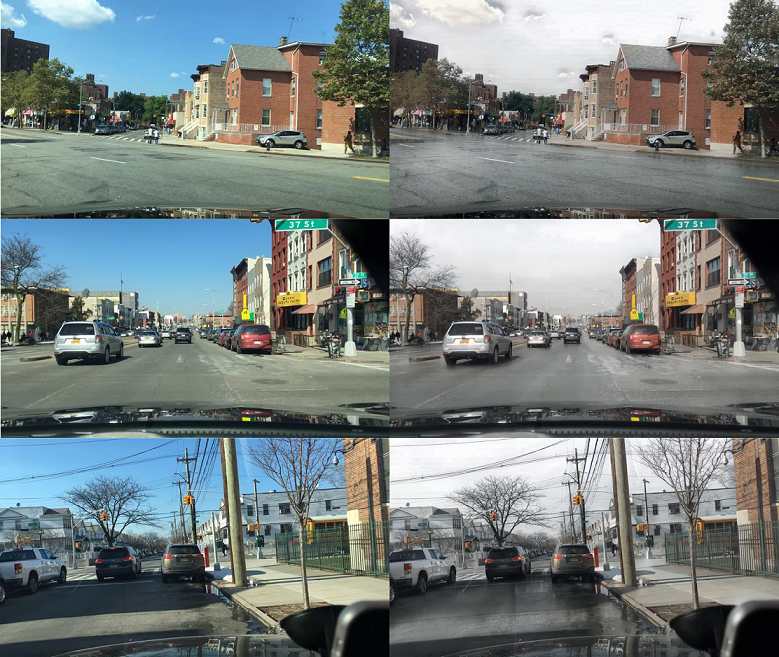}
\caption{The examples of generated images by the trained UNIT model, left: original clear images, right: generated rainy images.}
\label{Fig2}
\vspace{-0.6cm}
\end{figure}

To train the UNIT model that learns the mapping from clear images to rainy ones, we use the \textit{train clear} set that consists of clear-weather annotated images as the source domain. For the target rainy domain, we extract a sufficiently large number of images from the rainy videos of BDD100K data set. Subsequently, we apply images in the \textit{train clear} set to the trained UNIT model to generate rainy images. We refer to the images that are generated by the UNIT model as the \textit{train gen rainy} set. Examples of generated rainy images are shown in Figure \ref{Fig2}.

Eventually, we use the train gen rainy data set to train the detection methods. This is followed by using the \textit{test rain} data set to evaluate the average precision performance of the detection methods, which are now trained using the generated rainy set. We also calculate the mean average precision (mAP) as we have done for other approaches. Table \ref{mAP} shows the performance of detection methods that are trained using generated rainy images by the UNIT model.

\subsection{Domain adaptation} \label{DA}
Domain adaptation is another potentially viable framework that could be considered to address the major challenges that we have been highlighting in this tutorial regarding: (a) the salient mismatch between the two domains, clear and rainy weather conditions, and (b) the lack of annotated training data captured under rainy conditions. In particular, a domain adaptation framework \cite{chen2018domain} has been designed and developed specifically for Faster R-CNN due to the fact that it is among the most popular object detection approaches\footnote{At this point, we are not aware of other domain adaptation frameworks that have been designed and developed for YOLO. Consequently, given the tutorial nature of this paper, we only review domain adaptation that has already been developed for Faster R-CNN \cite{chen2018domain}.}.
The framework developed in \cite{chen2018domain} adapts deep learning based object detection to a target domain that is different from the training domain without requiring any annotations in the target domain. In particular, it employs the adversarial training strategy \cite{GAN} to learn robust features that are domain-invariant. In other words, it makes the distribution of features extracted from images in the two domain indistinguishable. 

The architecture for Domain Adaptive Faster R-CNN model \cite{chen2018domain} is shown in Figure \ref{Fig_faster-RCNN}. There are two levels of domain adaptation that are employed. First, an \textit{image-level} domain classifier is used. At this level, the global attributes, such as the image style, illumination, etc. of the input image are used to distinguish between the source and target domains. Thus, the (global) feature map resulting from the common CNN feature extractor of the Faster R-CNN detector is used as input toward the image-level domain classifier. Second, an \textit{instance-level} domain classifier is employed. This classifier uses the specific features associated with a particular region to distinguish between the two domains. Hence, the instance-level domain classifier uses the feature vector resulting from the fully connected layers (FCs) at the output of the RoI Pooling Layer of the Faster R-CNN detector. The two classifiers, the image- and instance-level ones, should naturally agree in terms of their binary classification decision regarding if the input image belongs to the source or target domain. Consequently, a \textit{consistency regularization} stage combines the output of the two classifiers to promote consistency between the two classifier outcomes.

While the two domain adaptation classifiers are optimized to differentiate between the source and target domains, the Faster R-CNN detector must be optimized such that it becomes domain-independent or \textit{domain-invariant}. In other words, the Faster R-CNN detector must detect objects regardless of the input image domain (clear or rainy). Hence, the feature map resulting from the Faster R-CNN feature extractor must be domain-invariant. To that end, this feature extractor should be trained and optimized to \textit{maximize} the domain classification error achieved by the domain adaptation stage. Thus, while both the image- and instance-level domain classifiers are designed to minimize the binary-classification error (between the source and target domains), the Faster R-CNN feature extractor is designed to maximize the same binary-classification error.

To achieve this contradictory objectives, a Gradient Reversal Layer (GRL) \cite{GRL} is employed. Thus, GRL is a bidirectional operator that is used to realize two different optimization objectives.  In the feed-forward direction, the GRL acts as an identity operator. This leads to the standard objective of minimizing the classification error when performing \textit{local} backpropagation within the domain adaptation network. On the other hand, for backpropagation toward the Faster R-CNN network, the GRL becomes a negative scalar. Hence, in this case, it leads to maximizing the binary-classification error; and this maximization promotes the generation of domain-invariant feature map by the Faster R-CNN feature extractor.

Consequently, for the purpose of this tutorial, we developed  and employed a domain adaptive faster R-CNN \cite{chen2018domain} under rainy conditions. To train this model, we prepare the training data to include two sets: \textit {source data}, which consists of images captured in clear weather (and this set includes data annotations in terms of bounding boxes coordinates and object categories), and \textit {target data}, which only consists of images captured under rainy conditions without any annotations. To validate the trained model using domain adaptation, we tested it using the \textit{test rainy} set.  The performance of the detection method (Faster R-CNN) that are trained by the domain adaptation approach is shown in the bottom row of Table \ref{mAP}.  

\subsection{Discussion}

\begin{figure*}[!t]
\centering
\includegraphics[width=1\linewidth]{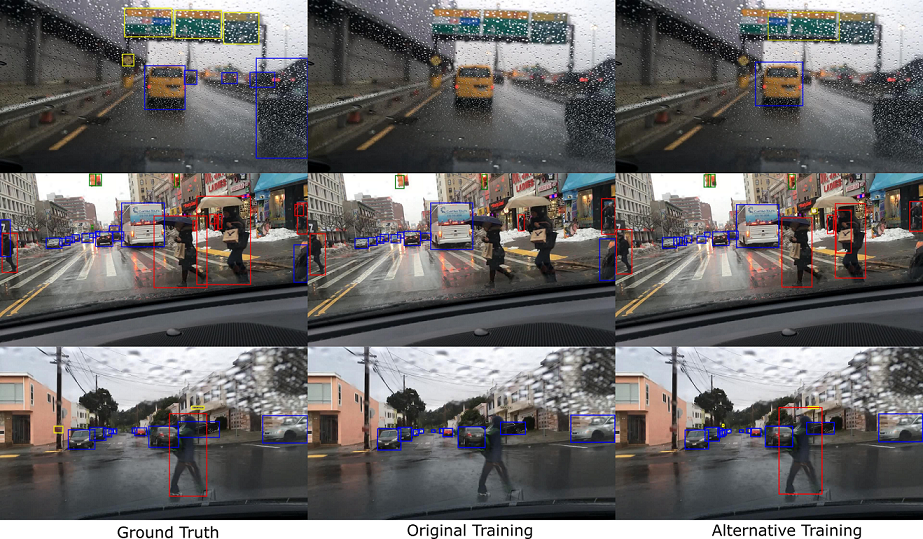}
\caption{The examples of detection results using alternative training approaches for Faster R-CNN and YOLO are shown in the rightmost column. The top-right image shows improvement in vehicle and traffic sign detection when generated images by I2IT (the UNIT model \cite{unit}) are used to train Faster R-CNN. The middle-right image shows improvement in pedestrian detection when domain adaptation \cite{chen2018domain} is used to train Faster R-CNN. The bottom-right image shows improvement in pedestrian detection when generated images by I2IT (the UNIT model \cite{unit}) are harnessed to train YOLO.}
\label{Fig_alternative}
\vspace{-0.6cm}
\end{figure*}

Based on the results in Table \ref{mAP}, we observe that while deraining algorithms degrade the average precision performance when tested on scenes distorted by natural rain, improvements can be achieved when employing image-to-image translation and domain adaptation as mitigating techniques. Different cases are presented in Figure \ref{Fig_alternative}. In terms of average precision, and as an example, rainy conditions degrade the pedestrian detection capabilities for YOLO by more than 5\% (from around 37\% to around 32\%); but by using image translation, the performance is improved to an average precision of more than 34\%, and consequently, narrowing the gap between clear and rainy conditions' performances.
Similarly, both image translation and domain adaptation improves the traffic-signs detection performance for Faster R-CNN. Furthermore, image translation seem to improve the vehicle detection performance under Faster R-CNN.


In other cases, for example, the traffic light detection performance under Faster R-CNN, the domain adaptation and image translation do not seem to perform well when tested on natural rainy images (even when using natural rainy images as the target domain for training these techniques). One potential factor for this poor performance in some of these cases is the fact that small objects, such as traffic lights, are quite challenging to detect to start with. This can be seen from the very low AP value, even in clear conditions, which is a mere 26\%. Naturally, the impact of raindrops or rain streaks on such small objects in the scene could be quite severe, to the extent that a mitigating technique might not be able to recover the salient features of these objects.

In summary, employing domain adaptation or generating rainy weather visuals using UNIT translation, and then using these visuals for training, seems to narrow the gap in performance due to the domain mismatch between clear and rainy weather conditions. This promising observation becomes especially clear when considering the disappointing performance of deraining algorithms. Nevertheless, it is also evident that there is still much room for improvement toward reaching the same level of performance in clear conditions. There are key challenges that need to be addressed, though, when designing any new mitigating techniques for closing the aforementioned gap. These challenges include the broad and diverse scenarios for rainy conditions, especially in driving environments.

\begin{figure*}[!t]
\centering
\includegraphics[width=1\linewidth]{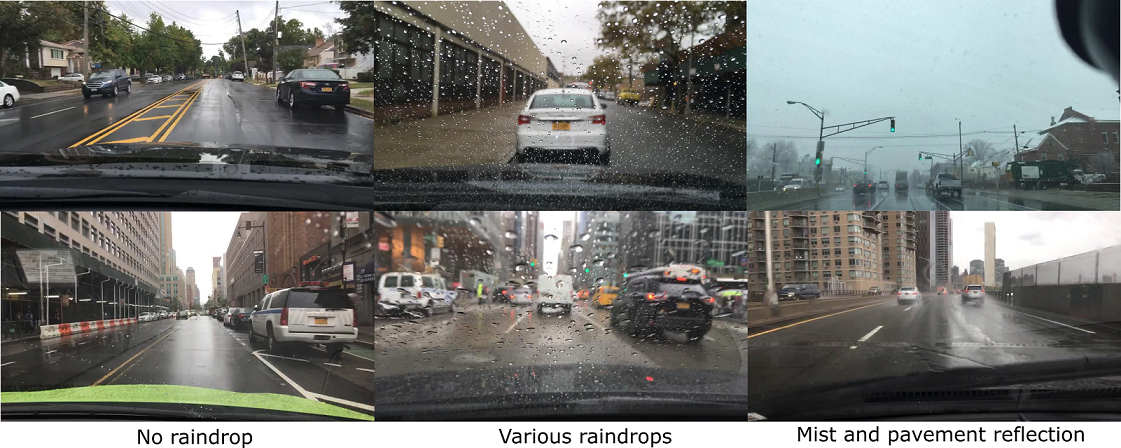}
\caption{The images from test-rainy set that illustrate several scenarios and effects of rainy weather for driving vehicles.}
\label{Fig5}
\vspace{-0.6cm}
\end{figure*}

These diverse cases and scenarios cannot be learned in a viable way by using state-of-the-art approaches. For example, raindrops have a wide range of possible appearances, and they come with various sizes and shapes, especially when falling on the windshield of a vehicle. Another factor is the influence of windshield wipers on altering the amount of rain, and even the shapes and sizes of raindrops, between wipe cycles. Other external factors include reflections from the surrounding wet pavement, mist in the air, and splash effects. Current state-of-the-art image translation techniques and domain adaptation are not robust enough to capture this wide variety of rain effects. Figure \ref{Fig5} provides images from the test rainy set that illustrate several rainy weather scenarios and effects for driving vehicles.   

\section{Conclusion} \label{sec_conclusion}
Besides outlining state-of-the-art frameworks for object de­­tection, deraining, I2IT, and domain adaptation, this tutorial highlighted crucial results and conclusions regarding current methods in terms of their performance in rainy weather conditions. In particular, we believe there is an overarching consistent message regarding the limitations of the surveyed techniques in handling and mitigating the impact of rain for visuals captured by moving vehicles. This consistent observation has serious implications for autonomous vehicles since the aforementioned limitations impact autonomous vehicles’ core safety capabilities. To address these issues, we recap some of our key findings and point out potential directions:
\begin{enumerate}[leftmargin=*]
    \item The lack of data, and especially annotated data, that captures the truly diverse nature of rainy conditions for moving vehicles is arguably the most critical and fundamental issue in this area. Major industry players are becoming more willing to tackle this problem and more open about addressing this issue publicly. Consequently, a few related efforts have just been announced and actually commenced by high-tech companies. These efforts are specifically dedicated to operating fleets of autonomous vehicles in challenging and diverse rainy weather conditions, explicitly for the sake of collecting data under these conditions \cite{Waymo-Florida}. After years of testing and millions of driven miles conducted primarily in favorable and clear weather, there is a salient admittance and willingness to divert important resources toward data collection in challenging weather conditions that will be encountered by autonomous vehicles.
    
    \item Despite the recent efforts to collect more diverse data, we believe that generative models could still play a crucial role in training object detection methods to be more robust and resilient in challenging conditions. In particular, we believe that novel and more advanced frameworks for UNIT could play a viable role for generating meaningful data for training. Due to the fact that these frameworks do not require annotated data, their underlying generative models could be useful in many ways. First, they could fill the gap that currently exists in terms of the lack of real annotated data in different weather conditions; hence, progress in terms of training and testing new object detection methods could be achieved by using these generative models. Second, even after a reasonable amount of annotated data captured in natural rainy conditions becomes available, the generative models could still play a pivotal role in both the basic training and coverage of diverse scenarios. In other words, UNIT models could always generate more data that can compliment real data, and this, on its own, could be quite helpful to further the basic training of object detection methods. Furthermore, despite the number of various rainy condition scenarios that real data actually represent, there will always be a need for capturing certain scenarios that are not included in a real data set. In that context, generative models could be used to produce data representing the scenarios that are missing from the real data sets, and hence they could increase the coverage and diversity of the cases that object detection methods can handle.
    \item There is a need for novel deep learning architectures and solutions that have adequate capacity for handling object detection under diverse conditions. Designing a neural network that performs quite well in one domain yet degrades in others is not a viable strategy for autonomous vehicles. In general, training the leading object detection architectures through a diverse set of data does not necessarily improve the performance of these architectures relative to their results when trained on a narrow domain of cases and scenarios. We believe that this issue represents an opportunity for researchers in the field to make key contributions.
\end{enumerate}

\section*{Acknowledgment}
This work has been supported by the Ford Motor Company under the Ford-MSU Alliance Program. The authors would like to acknowledge Ford's Advanced Research Engineers, Jonathan Diedrich and Mark Gehrke, for their invaluable input and contributions throughout the collaborative effort that resulted in this article.


\ifCLASSOPTIONcaptionsoff
  \newpage
\fi


\bibliography{ref.bib}
\bibliographystyle{IEEEtran}


%








\end{document}